\documentclass[conference]{IEEEtran}
\IEEEoverridecommandlockouts
\usepackage{cite}
\usepackage{amsmath,amssymb,amsfonts}
\usepackage{algorithmic}
\usepackage{graphicx}
\usepackage{textcomp}
\usepackage{xcolor}
\usepackage[hyphens]{url}

\def\BibTeX{{\rm B\kern-.05em{\sc i\kern-.025em b}\kern-.08em
    T\kern-.1667em\lower.7ex\hbox{E}\kern-.125emX}}
\begin{document}

\title{Do's and Don'ts for Human and Digital Worker Integration}

\author{
Vinod Muthusamy, Merve Unuvar, Hagen V\"olzer, Justin D. Weisz \\
IBM Research
}


\maketitle

\begin{abstract}
Robotic process automation (RPA) and its next evolutionary stage, \emph{intelligent process automation}, promise to drive improvements in efficiencies and process outcomes. However, how can business leaders evaluate \emph{how} to integrate intelligent automation into business processes? What is an appropriate division of labor between humans and machines? How should combined human-AI teams be evaluated? For RPA, often the human labor cost and the robotic labor cost are directly compared to make an automation decision. In this position paper, we argue for a broader view that incorporates the potential for multiple levels of autonomy and human involvement, as well as a wider range of metrics beyond productivity when integrating \emph{digital workers} into a business process.
\end{abstract}

\section{Introduction}
AI technologies are transforming how work is conducted in every industry. Today, AI systems are enabling businesses to personalize services, converse with customers, automate operations, optimize workflows, predict demand, and recommend next best actions. These AI-powered capabilities are enabling businesses to implement adaptive workflows in which humans and machines collaborate in novel ways. At least 61\% of business leaders agree that human-machine collaboration will help them achieve their strategic priorities faster and more efficiently~\cite{fourtane2019human}. In addition, it is anticipated that the automation of work activities enabled by AI systems will precipitate a shift in skills deemed valuable for human workers, toward those that involve logical reasoning and creative thinking, and away from those that involve collecting and processing data~\cite{manyika2017jobs}.

In robotic process automation (RPA), software ``robots'' perform repetitive, highly structured activities that have little variation. For example, RPA has been used to generate automatic responses to emails, automate the process of cancelling an airline ticket and issue a refund, as well as retrieve useful information from audit documents~\cite{boulton2018whatisrpa}. However, the utility of RPA is limited due to its reliance on simple logic and inability to reason or make decisions~\cite{greene2019overcoming, khalaf2017benefits}.

\emph{Intelligent process automation} represents the next phase of business automation in which AI capabilities are combined with the automative aspects of RPA~\cite{gadam2019digital} in a software agent called a \emph{digital worker.} By incorporating AI capabilities such as planning \& reasoning, image \& text recognition, and classification, digital workers will possess skills that correspond to human job roles and responsibilities~\cite{gadam2019digital}.

Although algorithmic advances in AI are being made at a rapid pace, less is known about how to incorporate digital workers into existing business processes to get the most benefit from AI technologies. There are many practical questions that a business leader or process owner must address when thinking through digital worker integration. In this position paper, we argue for an \emph{augmentative} rather than an \emph{automative} approach when incorporating digital workers into an existing business process. We present a framework for reasoning about digital worker integration that encompasses questions about whether to incorporate digital workers and into which parts of a business process, what patterns of interaction between human and digital workers should be adopted, and how the combined human + digital team should be evaluated.

\section{Related Work}
The topic of human vs. computer initiative has been discussed over the past 40 years~\cite{carbonell1970mixed, horvitz1999principles, liapis2016mixed, licklider1960man, seeber2020machines}, and several researchers in this space have developed frameworks that describe the interaction dynamics between humans and machines working together in a collaborative process. Biles~\cite{biles2002genjam} offered one analytic view that identified four simple paradigms of human-machine interaction: three patterns of \emph{unidirectional} human-AI dynamics, such as the human or machine generating alternatives and the other party selecting or modifying them; and one pattern of human-AI \emph{collaboration}, in which both party iteratively modify each others' creations. Parasuraman et al.~\cite{parasuraman2000model} approached the question of human-machine interaction from the perspective of \emph{levels of automation}: as the level of automation increases, machines shift from providing recommendations (levels 2-4) to executing on those recommendations with decreasing amounts of human oversight (levels 5-10). In a meta-analysis, Onnasch et al.~\cite{onnasch2014human} found that although increased automation had clear benefits for routine tasks, it was harmful for exceptional or failure cases and for maintaining situational awareness.

Both of these models implicitly identify the concept of \emph{initiative}: the responsibility for taking an action within a collaborative process. Mixed-initiative user interfaces~\cite{allen1999mixed} are those in which human and AI agents are independently able to make contributions to a shared task. Prior to this notion, there had been a debate within the HCI community as to which philosophy user interface research should follow: either, new user interface metaphors should be developed to further enhance the capabilities of human users to \emph{directly manipulate} objects, or intelligent agents should be developed that sense the intentions and goals of human operators and \emph{autonomously act} on their behalf. This debate is summarized by Horvitz~\cite{horvitz1999principles}, who argued for mixed-initiative user interfaces as an alternative area of inquiry.

Recently, Deterding et al. and Spoto offered a more detailed analysis of the microstructure of Biles' collaboration pattern through \emph{mixed-initiative creative interfaces}~\cite{deterding2017mixed, web:spoto}. In this framework, humans and machines engage in a wider range of activities: \emph{ideating}, \emph{constraining}, \emph{producing}, \emph{suggesting}, \emph{selecting}, \emph{assessing}, and \emph{adapting}. Significantly, each party is capable of performing each of these seven activities, although one party (human or machine) may be better suited for certain activities~\cite{stroganov2020role}.

The different interaction patterns discussed above are well-represented across a series of landmark systems developed for computer-assisted design, surveyed by Liapis et al.~\cite{liapis2016mixed}.
From their analysis of these tools, they proposed a number of heuristic questions for designers of mixed initiative tools, such as how one determines an appropriate division of labor between human and machine, and how conflicts between human and machine should be resolved. Our work is influenced by these crucial questions.

Another influential work in the human-machine interaction space was recently conducted by Seeber et al.~\cite{seeber2020machines}, who surveyed 65 collaboration researchers to develop a research agenda for exploring the potential risks and benefits of AI technologies in team collaborations. Their characterization of ``machine teammates'' is akin to our framing of ``digital workers.'' Several of the research questions posed by Seeber et al. also informed our framework of digital worker integration, including questions about how to determine what tasks should be executed by human, machine, or collaboratively (akin to Liapis et al.'s question of the division of labor), and how to identify problems that can benefit from the integration of human and machine knowledge. Similarly, Daugherty et al.~\cite{daugherty} recently examined the question of how business processes may change with the incorporation of AI technologies, and proposed a framework for how the role of human workers will shift as AI systems mature.

\section{A Framework for Digital Worker Integration}

We outline our framework as a set of questions that apply to different stages of the process development lifecycle. For each question, we provide examples of \emph{do's} that may lead to successful outcomes and \emph{don'ts} that may not (in bold).

\subsection{Whether to incorporate digital workers, and where?} \label{sec:whether}

\textbf{Do: Carefully consider the costs of ``hiring'' digital workers.} Good reasons for incorporating digital workers include improving human workers' productivity by freeing them from low-value or repetitive tasks, and making a business process more consistent or predictable by incorporating AI-based decision making. However, process owners may be tempted to incorporate AI into every possible step in a business process. We caution against this approach and instead recommend a careful cost-benefit analysis to understand the ROI from automation. Training AI models can be a costly process: obtaining and labeling training data, training state-of-the-art models on high-end compute systems, evaluating those models for performance, fairness, and robustness, and deploying those models into a production system all have costs that should be weighed against the potential benefits that come from digital workers (e.g. cost savings, additional revenue, improved customer or worker satisfaction). Second, AI models are susceptible to biases in the underlying data set, so care must be taken to ensure trained models are fair~\cite{bellamy2018ai}. Finally, to avoid negative consequences when a business process becomes \emph{too} automated\footnote{This is known as the \emph{paradox of automation}~\cite{bainbridge1983ironies}: as an automated system becomes more efficient, it simultaneously becomes more brittle, requiring human contributions to maintain.}, consideration needs to be given as to whether human workers will lose their skills when a digital worker takes over, and whether such de-skilling is tolerable.

\textbf{Do: Identify places in a workflow where a digital worker can help}. Within a business process, there may be certain stages or steps where human workers are struggling. These steps may be places at which digital workers can provide support to their human teammates. One way to identify these steps is to look for steps that have a high variance in performance metrics: some human workers perform a task well, and other human workers do not. This variance may be an indication that the task \emph{could} be performed better, since some human workers are already performing it well. At this stage, business process owners can evaluate whether and how a digital worker can help.

\textbf{Do: Scale confidently by scaling gradually.} It may be tempting for a process owner to fully incorporate a digital worker into the process and allow it to touch every case it comes across. We recommend a slower approach to build confidence in the digital worker: allow it to work on a small subset of active cases and have a human worker review its work before deciding whether it can be trusted to handle the full workload. In addition, if cases are able to be ranked based on their value or risk, consider applying the digital worker first to the low-value or low-risk cases before allowing it to affect higher-value or higher-risk cases. Even a digital worker that performs its work ``correctly'' can have unexpected side effects on downstream tasks in the process, on the morale of the human workers, or on customer satisfaction. A gradual approach affords space to identify and mitigate any potential harm the digital worker may cause.

\textbf{Don't: Assume that digital workers can do everything.} Digital workers are not a panacea and their limitations stem from limitations in AI technologies. For example, chatbots are often used to handle low-level support queries, diverting queries away from human workers~\cite{trips2017how}. This strategy may not work for workflows that require high levels of customer satisfaction (e.g. high profile clients). Additionally, some kinds of AI models may have limited accuracy; for example, state-of-the-art natural language classifiers are only able to achieve between 60-99\% accuracy~\cite{du2019explicit}, depending on the corpus. If high accuracy is a business requirement, ensuring that the AI model can provide it is a must before planning its incorporation into a business process.

\subsection{How should human and digital workers work together?} \label{sec:how}

\textbf{Do: Calibrate the initiative of digital workers.} Digital workers can act with varying levels of initiative, which determines how they will collaborate with their human co-workers~\cite{parasuraman2000model}. Digital workers that take no initiative only make recommendations to their human counterparts and take no actions themselves. Digital workers that possess full initiative act autonomously, never interacting with a human. These extreme ends of the spectrum may not provide enough aid (in the former case) or oversight (in the latter case). Intermediate levels of autonomy, such as when a digital worker makes a decision, notifies a human, and acts upon it within a time bound unless explicitly cancelled, may provide a desirable balance between aid and oversight. This decision should be guided by considering the efficacy of the digital worker, the risk of making a wrong decision, and the feasibility of compensating for mistakes.

\textbf{Do: Trust, but verify.} Guardrails are explicit bounds that are established to help manage risk that arises from the potential of making a mistake. For digital workers, guardrails that are based on business KPIs can be used to monitor when the quality of their decisions crosses a pre-determined risk threshold. Notifications can be triggered automatically when a digital worker performs an action that violates the guardrail, signaling the need for human intervention and possible compensation actions. In hybrid teams of human and digital workers, human workers should always possess the ability to override, revoke, or compensate the actions performed by a digital worker. In addition, when one of these actions is performed, human workers should be required to document their reasons for doing so in order to provide feedback and improve the performance of the system (discussed further below).

\textbf{Do: Empower and encourage human workers to improve their digital co-workers.} AI depends on data, both for the initial training and the subsequent improvement. Providing mechanisms for human workers to provide feedback on their digital co-workers is an important way to improve their utility and effectiveness. Feedback may be collected in a variety of ways. Quantitative feedback, such as ratings of how a digital worker performed on a specific task or corrections to how a digital worker classified an instance, may be directly used in the re-training process of the underlying AI model. Qualitative feedback, such as in-depth interviews with human co-workers, can be used to understand people's feelings toward the digital workers and how they have changed their work practices, as well as solicit novel ideas for how they can be improved. In the same way that business leaders often extol the virtues of a growth mindset~\cite{dweck2016having} in their human labor force, we assert that a similar mindset ought to be adopted for digital workers as well.

\textbf{Do: Make sure human workers understand their digital co-workers.} Trust is an integral component for hybrid human-digital worker teams to function successfully. If human teammates do not understand how their digital co-workers make decisions or why they made specific recommendations, they may not be apt to act upon them. Many techniques have been developed in the emerging field of Explainable AI (XAI)~\cite{adadi2018peeking,arrieta2020explainable,arya2019one} to provide explanations for how different types of AI models make decisions (known as global explanations) and for why an AI model made a decision in a specific instance (known as local explanations). In addition, knowing an AI system's confidence in its recommendation has been shown to help people calibrate their trust~\cite{zhang2020effect}. Recent work on improving the state of how AI systems are documented~\cite{arnold2019factsheets,hind2020experiences,shneiderman2020human} may also be instrumental in helping human workers understand and trust their digital co-workers.

\subsection{How to evaluate the combined human-digital workforce?}
\label{sec:performance}

\textbf{Don't: Directly compare the performance of human and digital workers.} Direct comparisons of human and digital workers using easy-to-measure metrics such as throughput or latency will be biased toward digital workers, who are able to complete their tasks faster and without repose. Managers should avoid such direct comparisons, and avoid exposing such simplistic performance metrics to their teams. Doing so may incentivize human workers to feel the need to compete against their digital co-workers by rushing their work, reducing the quality of their work, or avoiding taking on the higher-value or more complex cases that require human insight~\cite{muller_tyranny_2018}. Therefore, in order to evaluate the degree to which digital workers improve the quality of a business process, controlled A/B tests should be used in which a human control team's performance on a task is compared to the performance of a hybrid human+AI team on the same task. This way, the ``lift'' provided by digital workers on performance can be quantified.

\textbf{Do: Establish a holistic performance scorecard of the hybrid human-AI team.} It is important to capture a multitude of metrics for a business workflow, including performance (e.g. case completion time, length of the case backlog) and quality (e.g. accuracy or errors in decisions, revenue gain, or other types of ROI). However, other metrics are important as well, such as fairness in process outcomes, employee and customer satisfaction with the process, utilization of the digital workers, the extent to which feedback was provided on digital workers, and overriding behaviors. This last metric, which captures the number of instances in which a digital worker's activities or decisions were overridden by a human, is crucially important for monitoring the effectiveness of a digital worker. Too much overriding may signal problems with the quality of the underlying AI model, the consequences of which have been seen in many high-profile cases of AI model failure~\cite{yampolskiy2016artificial}. Conversely, not enough overriding may indicate that human workers are not providing enough oversight into a digital worker's activities.

\textbf{Do: Capture the interaction patterns between human and digital workers.}
Process analytics tools should make explicit which tasks are performed by human or digital workers, as well as identifying the collaboration patterns as touched on in Section~\ref{sec:how}. For example, understanding that a content extraction task (performed by a digital worker) and the verification task (performed by a human) are related (even if disjoint in the process), allows one to track mistakes by the digital worker, measure the work performed to correct the mistakes, and attribute any rework activities to these mistakes. These interactions become more subtle when digital workers are more autonomous with human workers only performing occasional oversight. We challenge the community to devise techniques to compute and visualize such metrics.

\textbf{Don't: Consider only quantitative metrics of performance when evaluating hybrid human+AI teams.} Qualitative metrics, such as how human workers feel toward their digital co-workers and how digital workers have changed aspects of a person's work life (e.g. workload, stress, job satisfaction, engagement) are also important indicators of the success of digital worker integration. A sole focus on performance outcomes will miss opportunities to receive feedback from human workers on how their digital co-workers have either improved or worsened the business process, and how the business process might be improved given the digital worker's capabilities. In this way, the philosophy of \emph{kaizen}~\cite{web:kaizen} -- a commitment to continuous improvement by all parties involved -- can be extended to include digital workers.

\section{Conclusion}
The advent of intelligent process automation will increasingly see humans and machines working together as a team to carry out knowledge work. Managing this hybrid workforce will require a principled approach that recognizes the value of both kinds of worker. In this position paper, we have outlined a number of recommendations for business leaders, managers, and business process owners who are thinking about digital worker integration. We address key questions about where to employ digital workers, how to structure their interactions with their human co-workers, and what metrics should be captured to evaluate the quality of these hybrid teams. Our insights touch on some of the opportunities and pitfalls in the lifecycle of managing human and digital workers, and we challenge the community to carry out further research in this area.

\bibliographystyle{splncs}
\bibliography{lib}

\end{document}